\documentclass[letterpaper, 10pt, conference]{ieeeconf}      

\IEEEoverridecommandlockouts                              
\usepackage{amsmath}    
\usepackage{amsfonts}
\usepackage{graphicx}   
\usepackage{subcaption}
\usepackage{epsfig} 
\usepackage{algorithmic}
\usepackage{color}
\usepackage[normalem]{ulem}
\usepackage{cancel}
\usepackage{amssymb}
\usepackage{color}
\usepackage[ruled,vlined,titlenotnumbered]{algorithm2e} 
\usepackage{cite}
\usepackage{float}

\newcommand{\pcset}{\mathcal{U}_p} 
\newcommand{\tcset}{\mathcal{U}_s} 
\newcommand{\tcfset}{\mathbb{U}_s} 
\newcommand{\dset}{\mathcal{D}}
\newcommand{\dfset}{\mathbb{D}}

\newcommand{\pset}{\mathcal{P}} 
\newcommand{\tset}{\mathcal{S}} 

\newcommand{\tvar}{t}
\newcommand{\thor}{T} 

\newcommand{\tstate}{s} 
\newcommand{\pstate}{p} 
\newcommand{\rstate}{r} 

\newcommand{\ttraj}{\xi_{\tdyn}} 
\newcommand{\rtraj}{\xi_\rdyn}

\newcommand{\senseDist}{m}

\newcommand{\tctrl}{u_s} 
\newcommand{\dstb}{d} 
\newcommand{\pctrl}{u_p} 

\newcommand{\tdyn}{f} 
\newcommand{\pdyn}{h} 
\newcommand{\rdyn}{g} 

\newcommand{\plannerfunc}{j}

\newcommand{\ptmat}{Q} 
\newcommand{\tpmat}{Q^T}

\newcommand{\errfunc}{l} 
\newcommand{\valfunc}{V} 

\newcommand{\deriv}{\nabla\valfunc} 

\newcommand{\dx}{\Delta x} 
\newcommand{\dt}{\Delta t} 

\newcommand{\obsSense}{\mathcal{O}_{sense}}
\newcommand{\obsAug}{\mathcal{O}_{aug}}

\newcommand{\TEB}{\mathcal B} 

\newtheorem{rem}{Remark}
\newtheorem{prop}{Proposition}

\title{\LARGE \bf FaSTrack: a Modular Framework for Fast and Guaranteed Safe Motion Planning}

\author{Sylvia L. Herbert*, Mo Chen*, SooJean Han, Somil Bansal, Jaime F. Fisac, and Claire J. Tomlin
\thanks{This research is supported by ONR under the Embedded Humans MURI (N00014-16-1-2206). The research of S. Herbert has received funding from the NSF GRFP and the UC Berkeley Chancellor's Fellowship Program.}
\thanks{* Both authors contributed equally to this work. All authors are with the Department of Electrical Engineering and Computer Sciences, University of California, Berkeley. \{sylvia.herbert, mochen72, soojean, somil, jfisac, tomlin\}@berkeley.edu}}

\begin{document}
\maketitle
\thispagestyle{empty}
\pagestyle{empty}

\begin{abstract}

Fast and safe navigation of dynamical systems through a priori unknown cluttered environments is vital to many applications of autonomous systems. However, trajectory planning for autonomous systems is computationally intensive, often requiring simplified dynamics that sacrifice safety and dynamic feasibility in order to plan efficiently. Conversely, safe trajectories can be computed using more sophisticated dynamic models, but this is typically too slow to be used for real-time planning. We propose a new algorithm FaSTrack: Fast and Safe Tracking for High Dimensional systems. A path or trajectory planner using simplified dynamics to plan quickly can be incorporated into the FaSTrack framework, which provides a safety controller for the vehicle along with a guaranteed tracking error bound. This bound captures all possible deviations due to high dimensional dynamics and external disturbances. Note that FaSTrack is modular and can be used with most current path or trajectory planners. We demonstrate this framework using a 10D nonlinear quadrotor model tracking a 3D path obtained from an RRT planner.

\end{abstract}

\section{Introduction}
\begin{figure}
	\centering
	\includegraphics[width=0.35\textwidth]{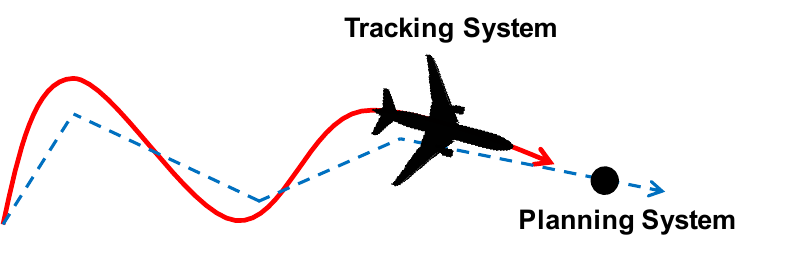}
	\caption{A planning system using a fast but simple model, followed by a tracking system using a dynamic model}
	\label{fig:chasing}
	\vspace{-.23in}
\end{figure}

 As unmanned aerial vehicles (UAVs) and other autonomous systems become more commonplace, it is essential that they be able to plan safe motion paths through crowded environments in real time. This is particularly crucial for navigating through environments that are \textit{a priori} unknown. However, for many common dynamical systems, accurate and robust path planning can be too computationally expensive to perform efficiently. In order to achieve real-time planning, many algorithms use highly simplified model dynamics or kinematics, resulting in a tracking error between the planned path and the true high-dimensional system. This concept is illustrated in Fig. \ref{fig:chasing}, where the path was planned using a simplified planning model, but the real vehicle cannot track this path exactly. In addition, external disturbances (e.g. wind) can be difficult to account for. Crucially, such tracking errors can lead to dangerous situations in which the planned path is safe, but the actual system trajectory enters unsafe regions.
 

We propose the modular tool FaSTrack: Fast and Safe Tracking, which models the navigation task as a sophisticated \textit{tracking system} that pursues a simplified \textit{planning system}. The tracking system accounts for complex system dynamics as well as bounded external disturbances, while the simple planning system enables the use of real-time planning algorithms. Offline, a precomputed pursuit-evasion game between the two systems is analyzed using Hamilton Jacobi (HJ) reachability analysis. This results in a \textit{tracking error function} that maps the initial relative state between the two systems to the \textit{tracking error bound}: the maximum possible relative distance that could occur over time. This tracking error bound can be thought of as a ``safety bubble" around the planning system that the tracking system is guaranteed to stay within. Because the tracking error is bounded in the relative state space, we can precompute and store a \textit{safety control function} that  maps the real-time relative state to the optimal safety control for the tracking system to ``catch" the planning system. It is important to note that the offline computations are \textit{independent} of the path planned in real-time; what matters are the relative states and dynamics between the systems, not the absolute state of the online path.

In the online computation, the autonomous system senses local obstacles, which are then augmented by the tracking error bound to ensure that no potentially unsafe paths can be computed. Next, a path or trajectory planner uses the simplified planning model to determine the next desired state. The tracking system then finds the relative state between itself and the next desired state. If this relative state is nearing the tracking error bound then it is plugged into the safety control function to find the instantaneous optimal safety control of the tracking system; otherwise, any controller may be used. In this sense, FaSTrack provides a \emph{least-restrictive} control law. This process is repeated until the navigation goal is reached.

Because we designed FaSTrack to be modular, it can be used with existing fast path or trajectory planners, enabling motion planning that is rapid, safe, and dynamically accurate. In this paper, we demonstrate this tool by computing the tracking error bound between a 10D quadrotor model affected by wind and a linear 3D kinematic model. Online, the simulated system travels through a static, windy environment with obstacles that are only known once they are within the limited sensing range of the vehicle. Combining this bound with a kinematic rapidly exploring random trees (RRT) fast path planner \cite{Gavin2013}, the system is able to safely plan and track a trajectory through the environment in real time.

\section{Related Work \label{sec:relatedwork}}
Motion planning is a very active area of research in the controls and robotics communities \cite{Hoy2015}.  In this section we will discuss past work on path planning, kinematic planning, and dynamic planning.  A major current challenge is to find an intersection of robust and real-time planning for general nonlinear systems. 

Sample-based planning methods like rapidly-exploring random trees (RRT) \cite{Kuffner2000}, probabilistic road maps (PRM) \cite{Kavraki1996}, fast marching tree (FMT) \cite{Janson2015}, and many others \cite{Richter2016, Karaman2011, Kobilarov2012} can find collision-free paths through known or partially known environments. While extremely effective in a number of use cases, these algorithms are not designed to be robust to model uncertainty or disturbances.

Motion planning for kinematic systems can also be accomplished through online trajectory optimization using methods such as TrajOpt \cite{Schulman2013} and CHOMP \cite{Ratliff2009}. These methods can work extremely well in many applications, but are generally challenging to implement in real time for nonlinear dynamic systems due to the computational load.

Model predictive control (MPC) has been a very successful method for dynamic trajectory optimization in both academia and industry \cite{Qin2003}.  However, combining speed, safety, and complex dynamics is a difficult balance to achieve. Using MPC for robotic and aircraft systems typically requires model reduction to take advantage of linear programming or mixed integer linear programming \cite{Vitus2008, Zeilinger2011, Richter2012}; robustness can also be achieved in linear systems \cite{Richards2006, DiCairano2016}. Nonlinear MPC is most often used on systems that evolve more slowly over time \cite{Diehl2002, Schildbach2016}, with active work to speed up computation \cite{Diehl2009, Neunert2016}. Adding robustness to nonlinear MPC is being explored through algorithms based on minimax formulations and tube MPCs that bound output trajectories with a tube around a nominal path (see \cite{Hoy2015} for references).


There are other methods of dynamic trajectory planning that manage to cleverly skirt the issue of solving for optimal trajectories online.  One such class of methods involve motion primitives \cite{Gillula2010, Dey2016}. Other methods include making use of safety funnels \cite{Majumdar2016}, or generating and choosing random trajectories at waypoints \cite{Kalakrishnan2011, Schwesinger2013}. The latter has been implemented successfully in many scenarios, but is risky in its reliance on finding randomly-generated safe trajectories. 

Recent work has considered using offline Hamilton-Jacobi analysis to guarantee tracking error bounds, which can then be used for robust trajectory planning \cite{Bansal2017}. A similar new approach, based on contraction theory and convex optimization, allows computation of offline error bounds that can then define safe tubes around a nominal dynamic trajectory computable online \cite{Singh2017}.

Finally, some online control techniques can be applied to trajectory tracking with constraint satisfaction. For control-affine systems in which a control barrier function can be identified, it is possible to guarantee forward invariance of the desired set through a state-dependent affine constraint on the control, which can be incorporated into an online optimization problem, and solved in real time \cite{Ames2014}. 

The work presented in this paper differs from the robust planning methods above because FaSTrack is designed to be modular and easy to use in conjunction with any path or trajectory planner. Additionally, FaSTrack can handle bounded external disturbances (e.g. wind) and work with both known and unknown environments with static obstacles.

\section{Problem Formulation \label{sec:formulation}}
In this paper we seek to simultaneously plan and track a trajectory (or path converted to a trajectory) online and in real time. The planning is done using a kinematic or dynamic planning model. The tracking is done by a tracking model representing the autonomous system. The environment may contain static obstacles that are either known a priori or can be observed by the system within a limited sensing range (see Section \ref{sec:online}). In this section we will define the tracking and planning models, as well as the goals of the paper.

\subsection{Tracking Model}
The tracking model is a representation of the autonomous system dynamics, and in general may be nonlinear and high-dimensional. Let $\tstate$ represent the state variables of the tracking model. The evolution of the dynamics satisfy the ordinary differential equation: 
\begin{equation}
\begin{aligned}
\label{eq:tdyn}
\frac{d\tstate}{d\tvar} = \dot{\tstate} = \tdyn(\tstate, \tctrl, \dstb), \tvar \in [0, \thor] \\
\tstate \in \tset, \tctrl \in \tcset, \dstb \in \dset
\end{aligned}
\end{equation}
We assume that the system dynamics $\tdyn : \tset\ \times\ \tcset \times \dset \rightarrow \tset$ are uniformly continuous, bounded, and Lipschitz continuous in $\tstate$ for fixed control $\tctrl$. The control function $\tctrl(\cdot)$ and disturbance function $\dstb(\cdot)$ are drawn from the following sets:
\begin{equation}
\begin{aligned}
\tctrl(\cdot) \in \tcfset(t) = \{\phi: [0, \thor] \rightarrow \tcset: \phi(\cdot) \text{ is measurable}\}\\
\dstb(\cdot) \in \dfset(t) = \{\phi: [0, \thor] \rightarrow \dset: \phi(\cdot) \text{ is measurable}\}
\end{aligned}
\end{equation}
where $\tcset, \dset$ are compact and $t\in[0, \thor]$ for some $T>0$. Under these assumptions there exists a unique trajectory solving (\ref{eq:tdyn}) for a given $\tctrl(\cdot) \in \tcset$ \cite{Coddington84}. The trajectories of (\ref{eq:tdyn}) that solve this ODE will be denoted as $\ttraj(\tvar; \tstate, \tvar_0, \tctrl(\cdot))$, where $\tvar_0,\tvar \in [0, \thor]$ and $\tvar_0 \leq \tvar$. These trajectories will satisfy the initial condition and the ODE (\ref{eq:tdyn}) almost everywhere:
\begin{equation}
\label{eq:fdyn_traj}
\begin{aligned}
\frac{d}{d\tvar}\ttraj(\tvar; \tstate, \tvar_0, \tctrl(\cdot)) &= \tdyn(\ttraj(\tvar; \tstate, \tvar_0, \tctrl(\cdot)), \tctrl(\tvar)) \\
\ttraj(\tvar; \tstate, \tvar, \tctrl(\cdot)) &= \tstate
\end{aligned}
\end{equation}

\subsection{Planning Model}
The planning model is used by the path or trajectory planner to solve for the desired path online. Kinematics or low-dimensional dynamics are typically used depending on the requirements of the planner. Let $\pstate$ represent the state variables of the planning model, with control $\pctrl$. The planning states $\pstate \in \pset$ are a subset of the tracking states $\tstate \in \tset$. The dynamics similarly satisfy the ordinary differential equation:
\begin{equation}
\begin{aligned}
\label{eq:pdyn}
\frac{d\pstate}{d\tvar} = \dot{\pstate} = \pdyn(\pstate, \pctrl), \tvar \in [0, \thor], \pstate \in \pset, \ \underline{\pctrl} \leq \pctrl \leq \overline{\pctrl}
\end{aligned}
\end{equation}
Note that the planning model does not involve a disturbance input. This is a key feature of FaSTrack: the treatment of disturbances is only necessary in the tracking model, which is modular with respect to any planning method, including those that do not account for disturbances.

\subsection{Goals of This Paper}
The goals of the paper are threefold:
\begin{enumerate}
	\item To provide a tool for precomputing functions (or look-up tables) to determine a guaranteed tracking error bound between tracking and planning models, and optimal safety controller for robust motion planning with nonlinear dynamic systems
	\item To develop a framework for easily implementing this tool with fast real-time path and trajectory planners.
	\item To demonstrate the tool and framework in an example using a high dimensional system
\end{enumerate}


\section{General Framework \label{sec:framework}}
\begin{figure}[h!]
  \centering
	\includegraphics[width=1\columnwidth]{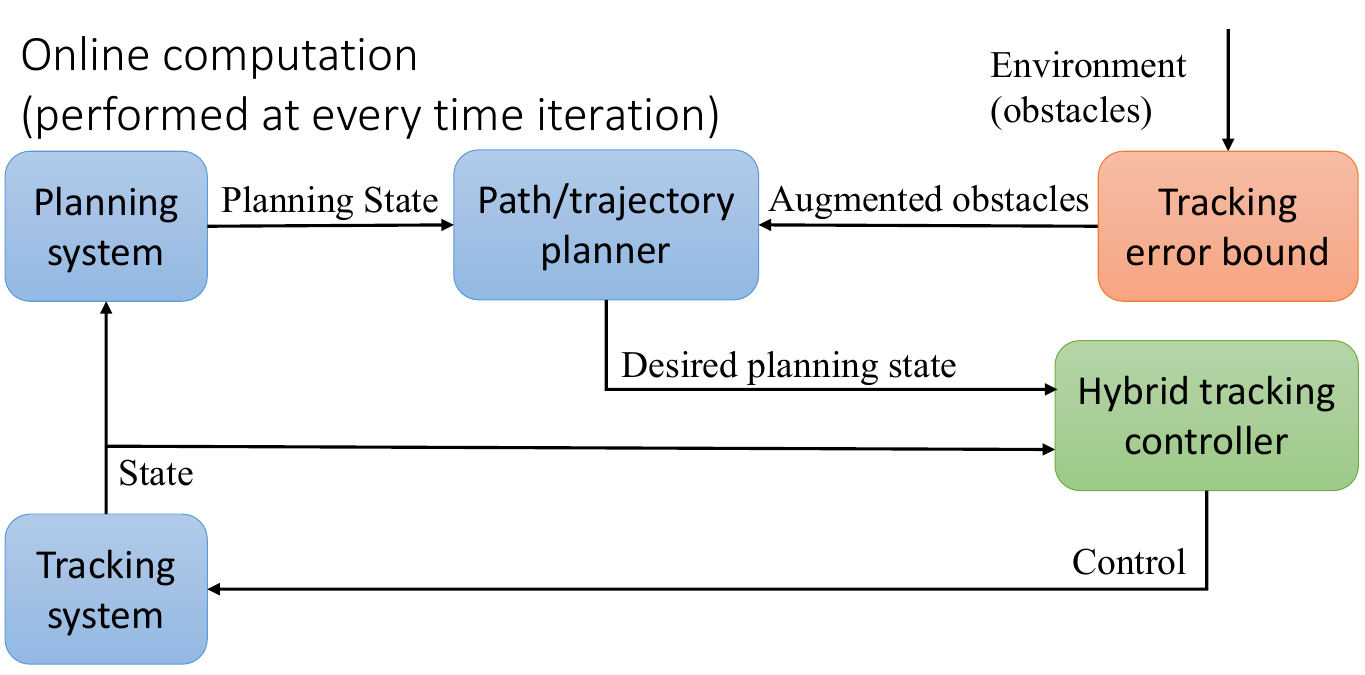}
	\caption{Online framework}
	\label{fig:fw_online}
	\vspace{-.1in}
\end{figure}

\begin{figure}[h!]
  \centering
	\includegraphics[width=0.9\columnwidth]{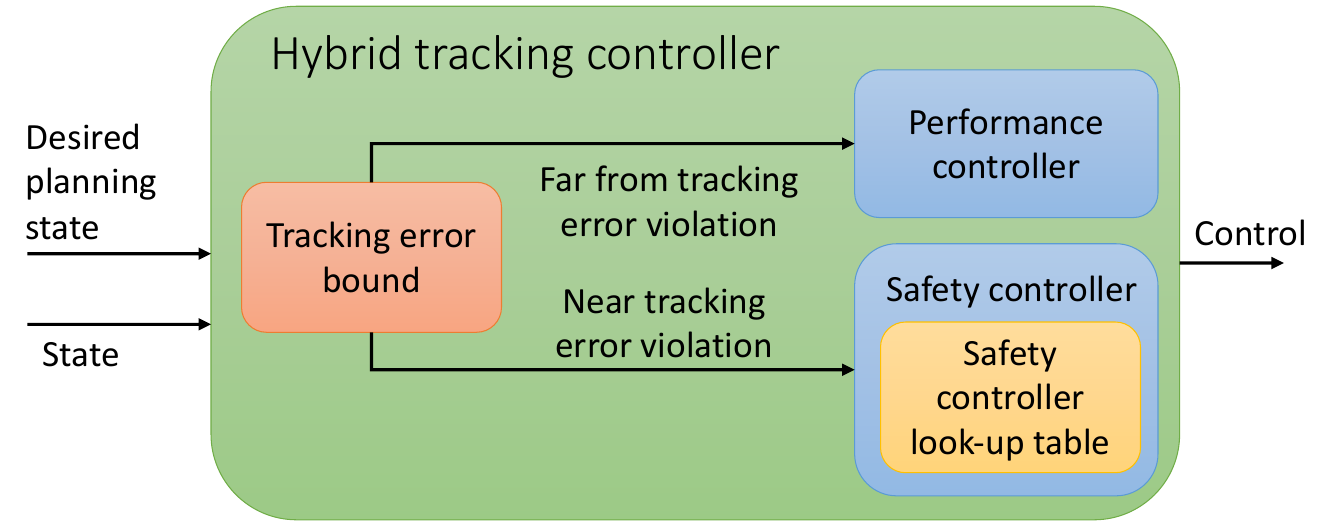}
	\caption{Hybrid controller}
	\label{fig:hybrid_ctrl}
	\vspace{-.1in}
\end{figure}

\begin{figure}[h!]
  \centering
	\includegraphics[width=0.9\columnwidth]{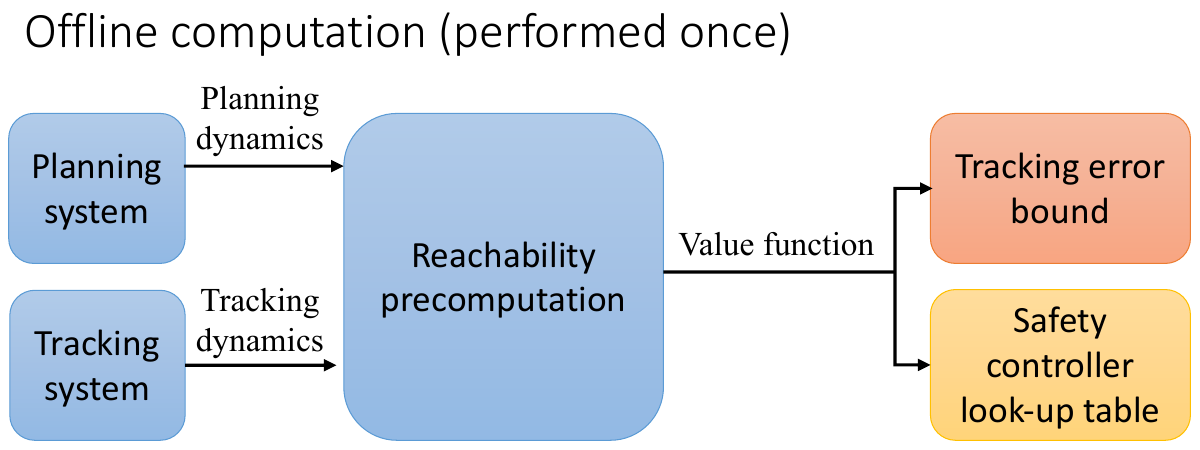}
	\caption{Offline framework}
	\label{fig:fw_offline}
\end{figure}

The overall framework of FaSTrack is summarized in Figs. \ref{fig:fw_online}, \ref{fig:hybrid_ctrl}, \ref{fig:fw_offline}. The online real-time framework is shown in Fig. \ref{fig:fw_online}. At the center of this framework is the path or trajectory planner; our framework is agnostic to the planner, so any may be used (e.g. MPC, RRT, neural networks). We will present an example using an RRT planner in Section \ref{sec:results}.

When executing the online framework, the first step is to sense obstacles in the environment, and then augment the sensed obstacles by a precomputed tracking error bound as described in Section \ref{sec:precomp}. This tracking error bound is a safety margin that guarantees robustness despite the worst-case disturbance. Augmenting the obstacles by this margin can be thought of as equivalent to wrapping the planning system with a ``safety bubble". These augmented obstacles are given as inputs to the planner along with the current state of the planning system. The planner then outputs the next desired state of the planning system. 

The tracking system is a model of the physical system (such as a quadrotor). The hybrid tracking controller block takes in the state of the tracking system as well as the desired state of the planning system. Based on the relative state between these two systems, the hybrid tracking controller outputs a control signal to the tracking system. The goal of this control is to make the tracking system track the desired planning state as closely as possible.

The hybrid tracking controller is expanded in Fig. \ref{fig:hybrid_ctrl} and consists of two controllers: a safety controller and a performance controller. In general, there may be multiple safety and performance controllers depending on various factors such as observed size of disturbances, but for simplicity we will just consider one safety and one performance controller in this paper. The safety controller consists of a function (or look-up table) computed offline via HJ reachability, and guarantees that the tracking error bound is not violated, \textit{despite the worst-case disturbance}. In addition, the table look-up operation is computationally inexpensive. When the system is close to violating the tracking error bound, the safety controller must be used to prevent the violation. On the other hand, when the system is far from violating the tracking error bound, any controller (such as one that minimizes fuel usage), can be used. This control is used to update the tracking system, which in turn updates the planning system, and the process repeats.

To determine both the tracking error bound and safety controller functions/look-up tables, an offline framework is used as shown in Fig. \ref{fig:fw_offline}. The planning and tracking system dynamics are plugged into an HJ reachability computation, which computes a value function that acts as the tracking error bound function/look-up table. The spatial gradients of the value function comprise the safety controller function/look-up table. These functions are independent of the online computations---they depend only on the \textit{relative} states and dynamics between the planning and tracking systems, not on the absolute states along the trajectory at execution time.

In the following sections we will first explain the precomputation steps taken in the offline framework. We will then walk through the online framework and provide a complete example.
\section{Offline Computation \label{sec:precomp}}
The offline computation begins with setting up a pursuit-evasion game \cite{Huang11, Chen17} between the tracking system and the planning system, which we then analyze using HJ reachability. In this game, the tracking system will try to ``capture" the planning system, while the planning system is doing everything it can to avoid capture. In reality the planner is typically not actively trying to avoid the tracking system, but this allows us to account for worst-case scenarios. If both systems are acting optimally in this way, we want to determine the largest relative distance that may occur over time. This distance is the maximum possible tracking error between the two systems.

\subsection{Relative Dynamics}
To determine the relative distance that may occur over time we must first define the relative states and dynamics between the tracking and planning models. The individual dynamics are defined in Section \ref{sec:formulation}, equations (\ref{eq:tdyn}) and (\ref{eq:pdyn}). The relative system is found by fixing the planning model to the origin and finding the dynamics of the tracking model relative to the planning model, as shown below.

\begin{equation}
\label{eq:rdyn}
\begin{aligned}
\rstate = \tstate - \ptmat\pstate, \qquad \dot\rstate = \rdyn(\rstate, \tctrl, \pctrl, \dstb)
\end{aligned}
\end{equation}

where $\ptmat$ matches the common states of $\tstate$ and $\pstate$ by augmenting the state space of the planning model (as shown in Section \ref{sec:results}). The relative states $\rstate$ now represent the tracking states relative to the planning states. Similarly, $\tpmat$ projects the state space of the tracking model onto the planning model: $\pstate = \tpmat(\tstate-\rstate)$. This will be used to update the planning model in the online algorithm.

\subsection{Formalizing the Pursuit-Evasion Game}
\begin{figure}
	\centering
	\includegraphics[width=0.45\textwidth]{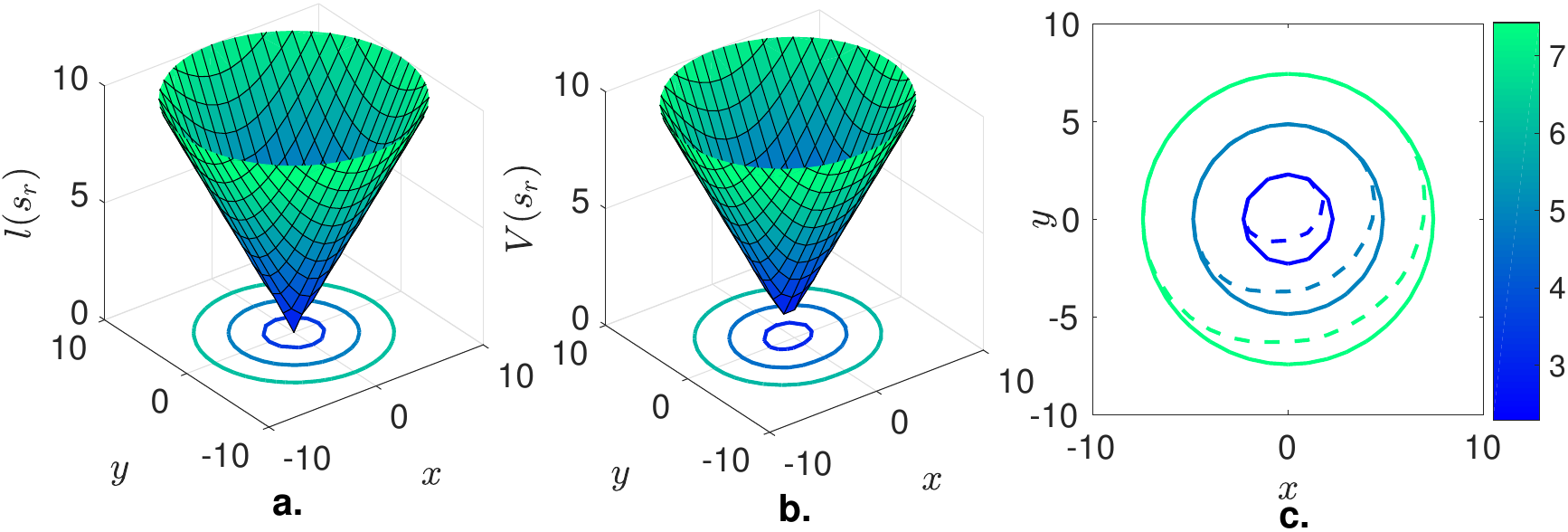}
	\caption{illustrative example of the precomputation steps for a 4D quadrotor model tracking a 2D kinematic planning model. All graphs are defined over a 2D slice of the 4D system. a) Cost function $\errfunc(\rstate)$ defined on relative states as distance to the origin, b) Value function $\valfunc(\rstate)$ computed using HJ reachability, c) Level sets of $\errfunc(\rstate)$ (solid) and $\valfunc(\rstate)$ (dashed). If the initial relative state is contained within the dashed set the system is guaranteed to remain within the corresponding solid set.}
	\label{fig:quad4D_example}
	\vspace{-.2in}
\end{figure}

Now that we have the relative dynamics between the two systems we must define a metric for the tracking error bound between these systems. We do this by defining an implicit surface function as a cost function $\errfunc(\rstate)$ in the new frame of reference. Because the metric we care about is distance to the origin (and thus distance to the planning system), this cost function can be as simple as distance in position space to the origin. An example can be seen in Fig. \ref{fig:quad4D_example}-a, where $\errfunc(\rstate)$ is defined for a 4D quadrotor model tracking a 2D kinematic planning model. The contour rings beneath the function represent varying level sets of the cost function. The tracking system will try to minimize this cost to reduce the relative distance, while the planning system will do the opposite.

Before constructing the differential game we must first determine the method each player must use for making decisions. We define a strategy for planning system as the mapping $\gamma_{\pstate} : \tcset \rightarrow \pcset$ that determines a control for the planning model based on the control of the planning model. We restrict $\gamma$ to draw from only non-anticipative strategies $\gamma_{\pstate} \in \Gamma_\pstate(t)$, as defined in \cite{Mitchell05}. We similarly define the disturbance strategy $\gamma_{\dstb}: \tcset \rightarrow \dset$, $\gamma_{\dstb} \in \Gamma_\dstb(t)$.

 We want to find the farthest distance (and thus highest cost) that this game will ever reach when both players are acting optimally. Therefore we want to find a mapping between the initial relative state of the system and the maximum possible cost achieved over the time horizon. This mapping is through our value function, defined as
 \begin{equation}
 \begin{aligned}
 \label{eq:valfunc}
 	&V(\rstate,\thor)= \sup_{\gamma_{\pstate} \in \Gamma_\pstate(t), \gamma_{\dstb} \in \Gamma_\dstb(t)} \inf_{\tctrl(\cdot) \in \tcfset(t)} \big\{\\
  &\qquad\qquad \max_{\tvar\in [0, \thor]} \errfunc\Big(\rtraj(\tvar; \rstate, 0, \tctrl(\cdot), \gamma_\pstate[\tctrl](\cdot), \gamma_\dstb[\tctrl](\cdot))\Big)\big\}
 	\end{aligned}
 \end{equation} 
 By implementing HJ reachability analysis we solve for this value function over the time horizon. If the control authority of the tracking system is powerful enough to always eventually reach the planning system, this value function will converge to an invariant solution for all time, i.e. $\valfunc_\infty(\rstate) := \lim_{\thor\rightarrow\infty} \valfunc(\rstate, \thor)$. An example of this converged value function is in Fig. \ref{fig:quad4D_example}-b. In the next section we will prove that the sub-level sets of this value function will map initial relative states to the guaranteed furthest possible tracking error over all time, as seen in Fig. \ref{fig:quad4D_example}-c.
 
In the context of the online framework, the value function $\valfunc_\infty(\rstate)$ is the tracking error bound function. The spatial gradients of the value function, $\deriv_\infty(\rstate)$, comprise the safety controller function (as described in Section \ref{sec:online}). When the framework is executed on a computer, these two functions are saved as look-up tables over a grid representing the state space of the relative system.
 
\subsection{Invariance of Converged Value Function}
 \begin{prop}
   \label{prop:main}
   Suppose that the value function converges, and define
      \begin{equation}
      \label{eq:conv_valfunc}
      \valfunc_\infty(\rstate) := \lim_{\thor\rightarrow\infty}\valfunc(\rstate, T)
      \end{equation}
   Then $\forall \tvar_1, \tvar_2$ with $\tvar_2 \ge \tvar_1$,
   \begin{equation}
   \label{eq:invariant}
   \valfunc_\infty(\rstate) \ge \valfunc_\infty\Big(\rtraj^*(\tvar_2; \rstate, \tvar_1)\Big), \text{where}
   \end{equation}
   \begin{equation}
   \begin{aligned}
   \rtraj^*(\tvar; \rstate, 0) := \rtraj(\tvar; \rstate, 0, \tctrl^*(\cdot), \pctrl^*(\cdot), \dstb^*(\cdot))) \\
   \end{aligned}
   \end{equation}
   \begin{equation}
   \begin{aligned}
   &\tctrl^*(\cdot) = \arg \inf_{\tctrl(\cdot)\in\tcfset(t)}\big\{\\
   & \qquad \max_{\tvar \in [0, \thor]} \errfunc(\rtraj(\tvar; \rstate, 0, \tctrl(\cdot), \pctrl^*(\cdot), \dstb^*(\cdot))) \big\}\\
    \end{aligned}
   \end{equation}
   \begin{equation}
   \begin{aligned}
   & \pctrl^*(\cdot) := \gamma_\pstate^*[\tctrl](\cdot) = \arg \sup_{\gamma_{\pstate} \in \Gamma_\pstate(t)} \inf_{\tctrl(\cdot) \in \tcfset(t)} \big\{ \\
   & \qquad \max_{t \in [0, \thor]} \errfunc(\rtraj(\tvar; \rstate, 0, \tctrl(\cdot), \gamma_\pstate[\tctrl](\cdot), \dstb^*(\cdot))) \big\} \\
    \end{aligned}
   \end{equation}
   \begin{equation}
   \begin{aligned}
   & \dstb^*(\cdot) = \arg \sup_{\gamma_{\dstb} \in \Gamma_\dstb(t)} \sup_{\gamma_{\pstate} \in \Gamma_\pstate(t)} \inf_{\tctrl(\cdot) \in \tcfset(t)} \big\{\\
   & \qquad \max_{\tvar \in [0, \thor]} \errfunc(\rtraj(\tvar; \rstate, 0, \tctrl(\cdot), \gamma_\pstate[\tctrl](\cdot), \gamma_\dstb[\tctrl](\cdot))) \big\}
   \end{aligned}
   \end{equation}
 \end{prop}

Proposition \ref{prop:main} proves that every level set of $\valfunc_\infty(\rstate)$ is invariant under the following conditions:
\begin{enumerate}
  \item The tracking system applies the optimal control which tries to track the planning system;
  \item The planning system applies (at worst) the optimal control that tries to escape from the tracking system; \label{ln:plan}
  \item The tracking system experiences (at worst) the optimal disturbance that tries to prevent successful tracking. \label{ln:dist}
\end{enumerate}
In practice, conditions \ref{ln:plan} and \ref{ln:dist} may not hold; the result of this is only advantageous to the tracking system and will make it easier to stay within its current level set of $\valfunc_\infty(\rstate)$, or to move to a smaller invariant level set of $\valfunc_\infty(\rstate)$. The smallest invariant level set corresponding to the value $\underline\valfunc := \min_{\rstate} \valfunc_\infty(\rstate)$ can be interpreted as the smallest possible tracking error of the system. The tracking error bound is given by\footnote{In practice, since $\valfunc_\infty$ is obtained numerically, we set $\TEB = \{\rstate: \valfunc_\infty(\rstate) \le \underline\valfunc + \epsilon\}$ for some suitably small $\epsilon>0$} the set $\TEB = \{\rstate: \valfunc_\infty(\rstate) \le \underline\valfunc\}$. This tracking error bound in the planner's frame of reference is given by:
\begin{equation} \label{eq:TEBp}
\TEB_\pstate(\tstate) = \{\pstate: \valfunc_\infty(\tstate-\ptmat\pstate) \le \underline\valfunc\}
\end{equation}
This is the tracking error bound that will be used in the online framework as shown in Fig. \ref{fig:fw_online}. Within this bound the tracking system may use any controller, but on the border of this bound the tracking system must use the safety optimal controller. We now prove Proposition \ref{prop:main}.

\begin{proof}
Without loss of generality, assume $\tvar_1=0$. By definition, we have
\begin{equation}
\begin{aligned}
\valfunc_\infty(\rstate) & = \lim_{\thor\rightarrow\infty}\max_{\tvar \in [0, \thor]} \errfunc(\rtraj^*(\tvar; \rstate, 0))\\
\end{aligned}
\end{equation}
By time-invariance, for some $\tvar_2 > 0$,
\begin{equation}
\label{eq:valfunc_ineq}
  \begin{aligned}
\valfunc_\infty(\rstate) &= \lim_{\thor\rightarrow\infty}\max_{\tvar \in [-\tvar_2, \thor]} \errfunc(\rtraj^*(\tvar; \rstate, -\tvar_2)) \\
&\ge \lim_{\thor\rightarrow\infty}\max_{\tvar \in [0, \thor]} \errfunc(\rtraj^*(\tvar; \rstate, -\tvar_2)) 
  \end{aligned}
\end{equation}  
\noindent where the sub-interval $[-\tvar_2, 0)$ has been removed in the last line. Next, by time invariance again, we  have
\begin{equation}
\begin{aligned}
\rtraj^*(\tvar; \rstate, -\tau) &= \rtraj^*(\tvar; \rtraj^*(0; \rstate, -\tvar_2), 0) \\
&= \rtraj^*(\tvar; \rtraj^*(\tvar_2; \rstate, 0), 0)
\end{aligned}
\end{equation}
Now, \eqref{eq:valfunc_ineq} implies
\begin{equation}
\begin{aligned}
\valfunc_\infty(\rstate) &\ge \lim_{\thor\rightarrow\infty}\max_{\tvar \in [0, \thor]} \errfunc(\rtraj^*(\tvar; \rtraj^*(\tvar_2; \rstate, 0), 0)) \\
&= \valfunc_\infty(\rtraj^*(\tvar_2; \rstate, 0))
\end{aligned}
\end{equation} 
\end{proof} 
 \begin{rem} 
   Proposition \ref{prop:main} is very similar to well-known results in differential game theory with a slightly different cost function \cite{Akametalu2014}, and has been utilized in the context of using the subzero level set of $\valfunc_\infty$ as a backward reachable set for tasks such as collision avoidance or reach-avoid games \cite{Mitchell05}. In our work we do not assign special meaning to any particular level set, and instead consider all level sets at the same time. This effectively allows us to perform solve many simultaneous reachability problems in a single computation, thereby removing the need to check whether resulting invariant sets are empty, as was done in \cite{Bansal2017}.
 \end{rem}

\section{Online Computation \label{sec:online}}
Algorithm \ref{alg:algOnline} describes the online computation. The inputs are the tracking error function $\valfunc_\infty(\rstate)$ and the safety control look-up function $\deriv_\infty(\rstate)$. Note that when discretized on a computer these functions will be look-up tables; practical issues arising from sampled data control can be handled using methods such as \cite{Mitchell2012, Mitchell13, Dabadie2014} and are not the focus of our paper.

Lines \ref{ln:Istart}-\ref{ln:Iend} initialize the computation by setting the planning and tracking model states (and therefore the relative state) to zero. The tracking error bound in the planning frame of reference is computed using (\ref{eq:TEBp}). Note that by initializing the relative state to be zero we can use the smallest possible invariant $\TEB_\pstate$ for the entire online computation. 
\begin{algorithm}	
	\caption{Online Trajectory Planning}
	\label{alg:algOnline}
	\begin{algorithmic}[1]
		\STATE \textbf{Initialization}: \label{ln:Istart}
		\STATE $\pstate = \tstate = \rstate = 0$
		\STATE $\TEB_\pstate(0) = \{\pstate: \valfunc_\infty(0) \le \underline\valfunc\}$ \label{ln:Iend}
		
		\WHILE{planning goal is not reached}
		\STATE \textbf{Tracking Error Bound Block}: \label{ln:obsStart}
		\STATE $\obsAug \leftarrow \obsSense + \TEB_\pstate(0)$ \label{ln:obsEnd}
		
		\STATE \textbf{Path Planner Block}:\label{ln:plannerStart}
		\STATE $\pstate_{next} \leftarrow \plannerfunc(\pstate, \obsAug)$\label{ln:plannerEnd}
		
		\STATE \textbf{Hybrid Tracking Controller Block}:\label{ln:controllerStart}
		\STATE $\rstate_{next} = \tstate - \ptmat\pstate_{next}$
		
		\IF{$\rstate_{next}$ is on boundary $\TEB_\pstate(0)$} 
		\STATE {use safety controller: $\tctrl \leftarrow \tctrl^*$ in \eqref{eq:opt_ctrl}}
		\ELSE \STATE{use performance controller: } 
          \STATE{$\tctrl \leftarrow$ desired controller} \ENDIF \label{ln:controllerEnd}
		
		\STATE \textbf{Tracking Model Block}: \label{ln:trackingStart}
		\STATE apply control $\tctrl$ to vehicle for a time step of $\dt$, save next state as $\tstate_{next}$ \label{ln:trackingEnd}
		
		\STATE \textbf{Planning Model Block}:\label{ln:planningStart}
		\STATE $\pstate = \tpmat\tstate_{next}$
		\STATE check if $\pstate$ is at planning goal
		\STATE reset states $\tstate = \tstate_{next}, \rstate = 0$ \label{ln:planningEnd}
		\ENDWHILE
	\end{algorithmic}
\end{algorithm}
The tracking error bound block is shown on lines \ref{ln:obsStart}-\ref{ln:obsEnd}. The sensor detects obstacles $\obsSense$ within the sensing distance around the vehicle. The sensed obstacles are augmented by $\TEB_\pstate(0)$ using the Minkowski sum. This is done to ensure that no unsafe path can be generated\footnote{The minimum allowable sensing distance is $\senseDist = 2\TEB_\pstate(0) + \dx$, where $\dx$ is the largest step in space that the planner can make in one time step.}.


 The path planner block (lines \ref{ln:plannerStart}-\ref{ln:plannerEnd}) takes in the planning model state $\pstate$ and the augmented obstacles $\obsAug$, and outputs the next state of the planning system $\pstate_{next}$. The hybrid tracking controller block (lines \ref{ln:controllerStart}-\ref{ln:controllerEnd}) first computes the updated relative state $\rstate_{next}$. If the $\rstate_{next}$ is on the tracking bound $\TEB_\pstate(0)$, the safety controller must be used to remain within the safe bound. The safety control is given by:
\begin{equation}
  \label{eq:opt_ctrl}
	\tctrl^* = \arg\min_{\tctrl\in\tcset} \max_{\pctrl\in\pcset, \dstb\in\dset} \nabla\valfunc(\rstate_{next}) \cdot \rdyn(\rstate_{next},\tctrl,\pctrl,\dstb)
\end{equation}
For many practical systems (such as control affine systems), this minimization can be found extremely quickly.

If the relative state is not on the tracking boundary, a performance controller may be used. For the example in Section \ref{sec:results} the safety and performance controllers are identical, but in general this performance controller can suit the needs of the individual applications.

The control $\tctrl^*$ is then applied to the physical system in the tracking block (lines \ref{ln:trackingStart}-\ref{ln:trackingEnd}) for a time period of $\dt$. The next state is saved as $\tstate_{next}$. This then updates the planning model state in the planning model block (lines \ref{ln:planningStart}-\ref{ln:planningEnd}). We repeat this process until the planning goal has been reached.

\section{10D Quadrotor RRT Example \label{sec:results}}
\begin{figure*}
	\centering
	\includegraphics[width=0.7\textwidth]{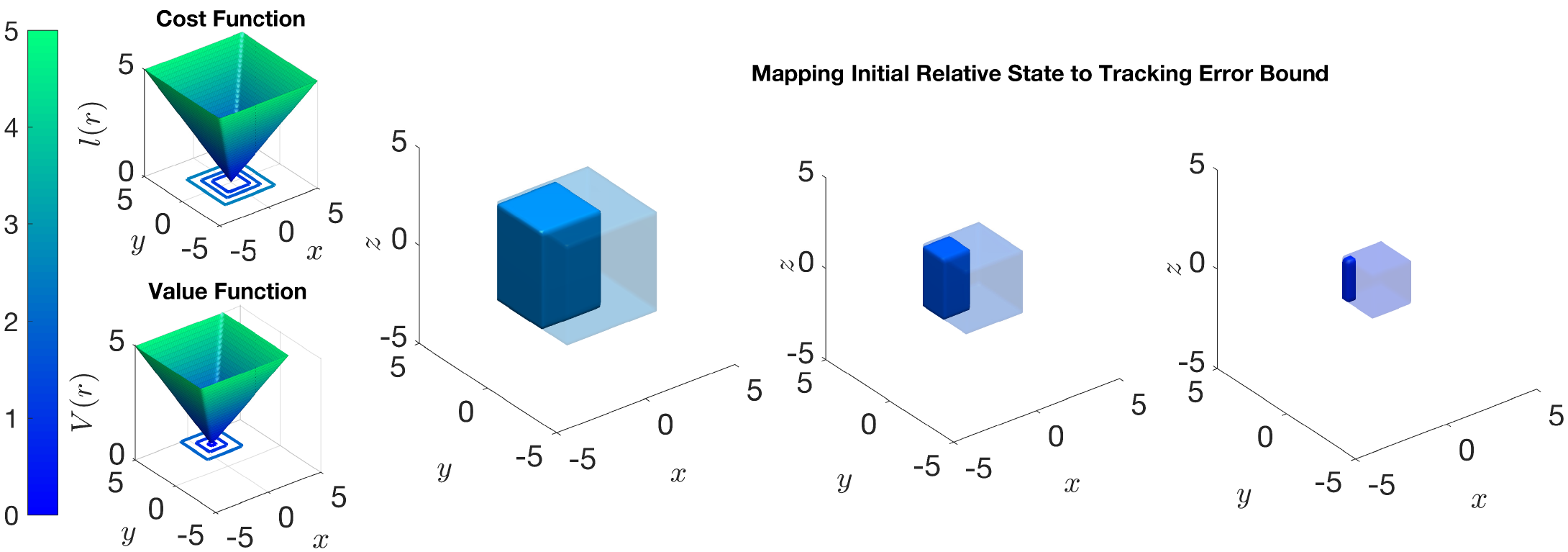}
	\caption{On the left are the cost and value functions over a 2D slice of the 10D relative state space, with contour lines showing three level sets of these functions. On the right are 3D projections of these level sets at the same slice $(v_{x},v_{y},v_{z})=[1, -1, 1]$ m/s, $(\theta_{x},\omega_{x},\theta_{y},\omega_{y})=0$. The solid boxes show initial relative states, and the transparent boxes show the corresponding tracking error bound. In practice we set the initial relative states to 0 to find the smallest invariant tracking error bound.}
	\label{fig:quad10D_example}
		\vspace{-.2in}
	\end{figure*} 
\begin{figure}
	\centering
	\includegraphics[width=0.25\textwidth]{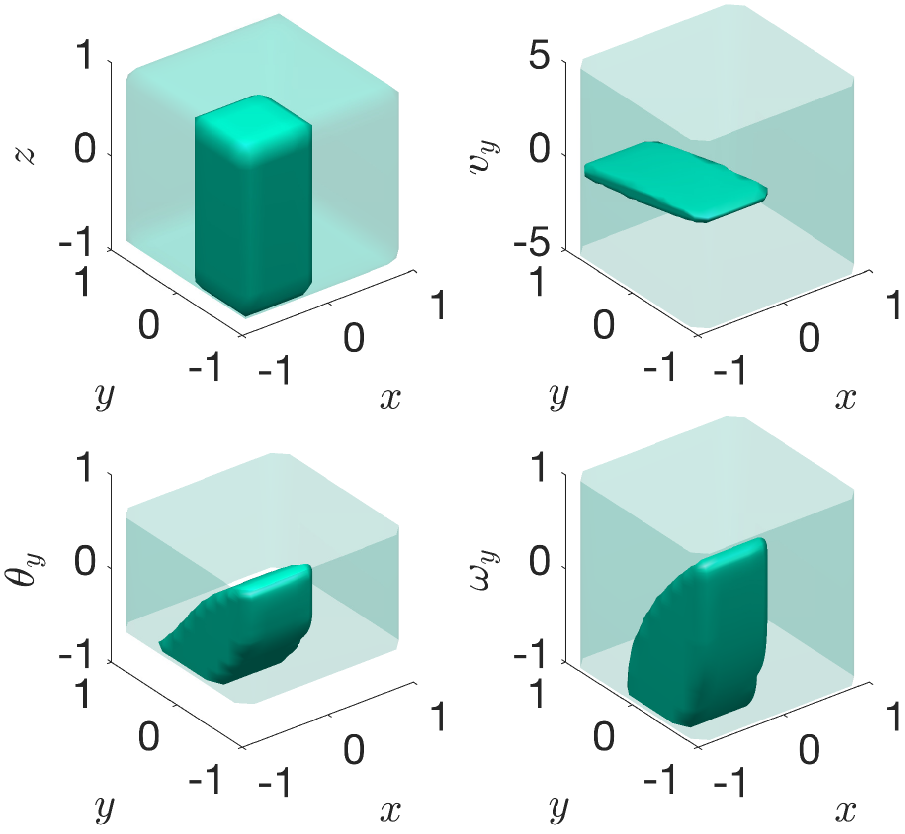}
	\caption{Various 3D slices of the 10D relative states (solid) and the corresponding tracking error bound (transparent)}
	\label{fig:quad10D_example_slices}
		\vspace{-.21in}
\end{figure} 
We demonstrate this framework with a 10D near-hover quadrotor developed in \cite{Bouffard12} tracking a 3D point source path generated by an RRT planner \cite{Gavin2013}. First we perform the offline computations to acquire the tracking error bound and safety controller look-up tables. Next we set up the RRT to convert paths to simple 3D trajectories. Finally we implement the online framework to navigate the 10D system through a 3D environment with static obstacles.

\subsection{Precomputation of 10D-3D system}
First we define the 10D dynamics of the tracking quadrotor and the 3D dynamics of a holonomic vehicle:
\begin{equation}
\label{eq:Quad10D_dyn}
\begin{aligned}
\begin{array}{c}
\left[
\begin{array}{c}
\dot{x}\\
\dot{v_x}\\
\dot{\theta_x}\\
\dot\omega_x\\
\dot{y}\\
\dot{v_y}\\
\dot{\theta_y}\\
\dot\omega_y\\
\dot{z}\\
\dot{v_z}
\end{array}
\right]
=
\left[
\begin{array}{c}
v_x + d_x\\
g \tan \theta_x\\
-d_1 \theta_x + \omega_x\\
-d_0 \theta_x + n_0 a_x\\
v_y + d_y\\
g \tan \theta_y\\
-d_1 \theta_y + \omega_y\\
-d_0 \theta_y + n_0 a_y\\
v_z + d_z\\
k_T a_z - g
\end{array}
\right]
\left[
\begin{array}{c}
\dot{x}\\
\dot{y}\\
\dot{z}\\
\end{array}
\right] 
=
\left[
\begin{array}{c}
b_x\\
b_y\\
b_z \\
\end{array}
\right]
\end{array}\\
\end{aligned}
\end{equation}
where states $(x, y, z)$ denote the position, $(v_x, v_y, v_z)$ denote the velocity, $(\theta_x, \theta_y)$ denote the pitch and roll, and $(\omega_x, \omega_y)$ denote the pitch and roll rates. The controls of the 10D system are $(a_x, a_y, a_z)$, where $a_x$ and $a_y$ represent the desired pitch and roll angle, and $a_z$ represents the vertical thrust. The 3D system controls are $(b_x, b_y, b_z)$, and represent the velocity in each positional dimension. The disturbances in the 10D system $(d_x, d_y, d_z)$ are caused by wind, which acts on the velocity in each dimension. Note that the states of the 3D dynamics are a subset of the 10D state space; the matrix Q used in the online computation matches the position states of both systems. Next the relative dynamics between the two systems is defined using (\ref{eq:rdyn}):
\begin{equation}
\label{eq:Quad10DRel_dyn}
\begin{aligned}
\begin{array}{c}
\left[
\begin{array}{c}
\dot{x_r}\\
\dot{v_{x}}\\
\dot{\theta_{x}}\\
\dot\omega_{x}\\
\dot{y_r}\\
\dot{v_{y}}\\
\dot{\theta_{y}}\\
\dot\omega_{y}\\
\dot{z_r}\\
\dot{v_{z}}
\end{array}
\right]
=
\left[
\begin{array}{c}
v_x - b_x + d_x\\
g \tan \theta_x\\
-d_1 \theta_x + \omega_x\\
-d_0 \theta_x + n_0 a_x\\
v_y - b_y + d_y\\
g \tan \theta_y\\
-d_1 \theta_y + \omega_y\\
-d_0 \theta_y + n_0 a_y\\
v_z - b_z + d_z\\
k_T a_z - g
\end{array}
\right]
\end{array}\\
\end{aligned}
\end{equation}
The values for parameters $d_0,d_1,n_0,k_T,g$ used were: $d_0=10,d_1=8,n_0=10,k_T=0.91,g=9.81$. The 10D control bounds were $|a_x|,|a_y|\leq10$ degrees, $0\leq a_z\leq 1.5g$ m/s$^{2}$. The 3D control bounds were $|b_x|,|b_y|,|b_z|\leq0.5$ m/s. The disturbance bounds were $|d_x|,|d_y|,|d_z|\leq0.1$ m/s. 

Next we follow the setup in section \ref{sec:precomp} to create a cost function, which we then evaluate using HJ reachability until convergence to produce the invariant value function as in (\ref{eq:valfunc}). Historically this 10D nonlinear relative system would be intractable for HJ reachability analysis, but using new methods in \cite{Chen2016DecouplingExact, Chen2016DecouplingJournal} we can decompose this system into 3 subsystems (for each positional dimension). Doing this also requires decomposing the cost function; therefore we represent the cost function as a 1-norm instead of a 2-norm. This cost function as well as the resulting value function can be seen projected onto the $x,y$ dimensions in Fig. \ref{fig:quad10D_example}.

Fig. \ref{fig:quad10D_example} also shows 3D positional projections of the mapping between initial relative state to maximum potential relative distance over all time (i.e. tracking error bound). If the real system starts exactly at the origin in relative coordinates, its tracking error bound will be a box of $\underline\valfunc = 0.81$ m in each direction. Slices of the 3D set and corresponding tracking error bounds are also shown in Fig. \ref{fig:quad10D_example_slices}. We save the look-up tables of the value function (i.e. the tracking error function) and its spatial gradients (i.e. the safety controller function).

\subsection{Online Planning with RRT and Sensing}
Our precomputed value function can serve as a tracking error bound, and its gradients form a look-up table for the optimal tracking controller. These can be combined with any planning algorithm such as MPC, RRT, or neural-network-based planners in a modular way. 

To demonstrate the combination of fast planning and provably robust tracking, we used a simple multi-tree RRT planner implemented in MATLAB modified from \cite{Gavin2013}. We assigned a speed of $0.5$ m/s to the piecewise linear paths obtained from the RRT planner, so that the planning model is as given in \eqref{eq:Quad10D_dyn}. Besides planning a path to the goal, the quadrotor must also sense obstacles in the vicinity. For illustration, we chose a simple virtual sensor that reveals obstacles within a range of 2 m in the $x$, $y$, or $z$ directions.

Once an obstacle is sensed, the RRT planner replans while taking into account all obstacles that have been sensed so far. To ensure that the quadrotor does not collide with the obstacles despite error in tracking, planning is done with respect to augmented obstacles that are ``expanded'' from the sensed obstacles by $\underline\valfunc$ in the $x$, $y$, and $z$ directions.

On an unoptimized MATLAB implementation on a desktop computer with a Core i7-2600K CPU, each iteration took approximately $25$ ms on average. Most of this time is spent on planning: obtaining the tracking controller took approximately $5$ ms per iteration on average. The frequency of control was once every $100$ ms.

\begin{figure}
	\centering
	\begin{subfigure}[t]{0.49\columnwidth} \label{subfig:sim_4}
		\includegraphics[width=\columnwidth]{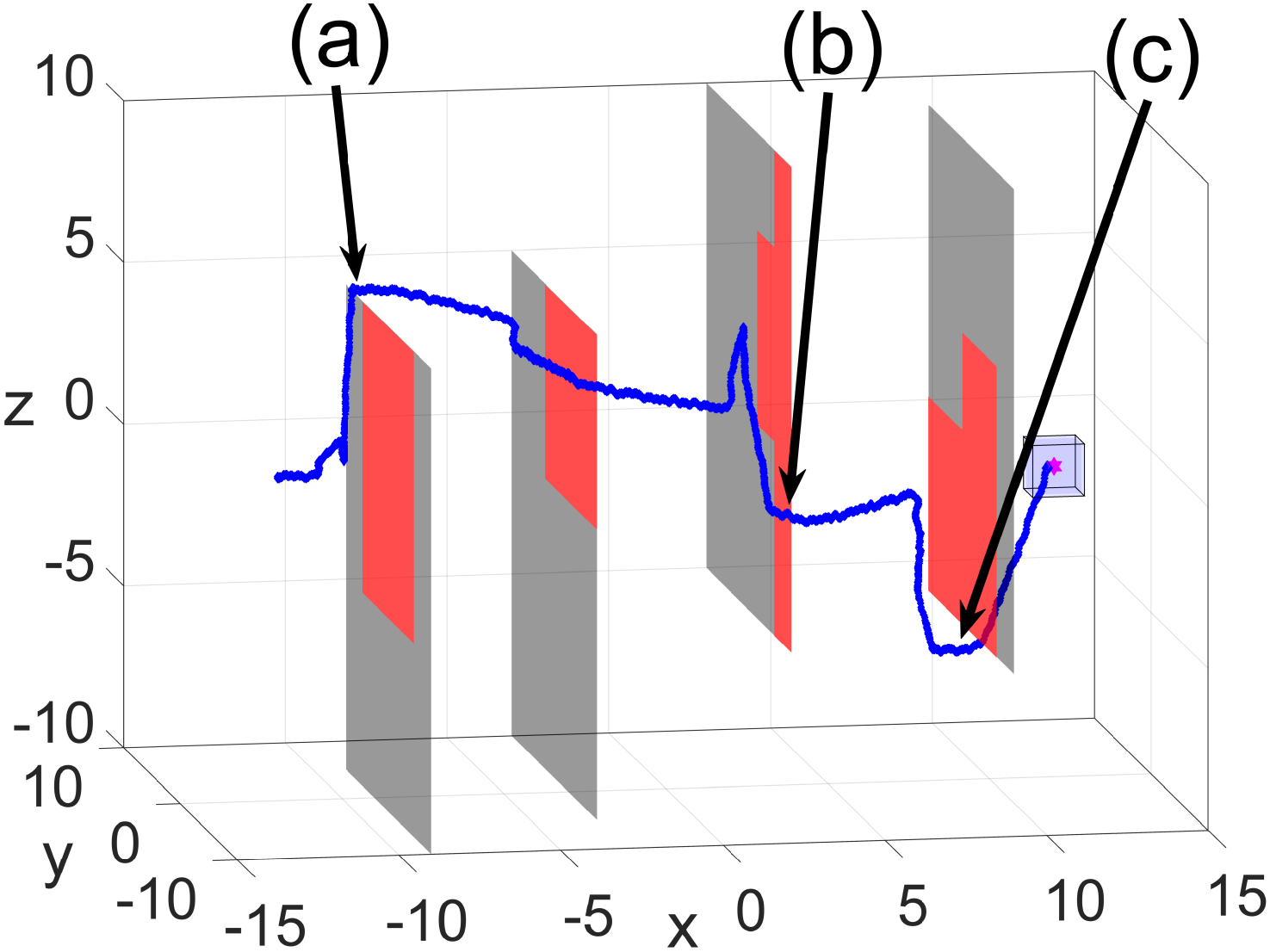}
	\end{subfigure}  
	\begin{subfigure}[t]{0.49\columnwidth} \label{subfig:sim_1}
		\includegraphics[width=\columnwidth]{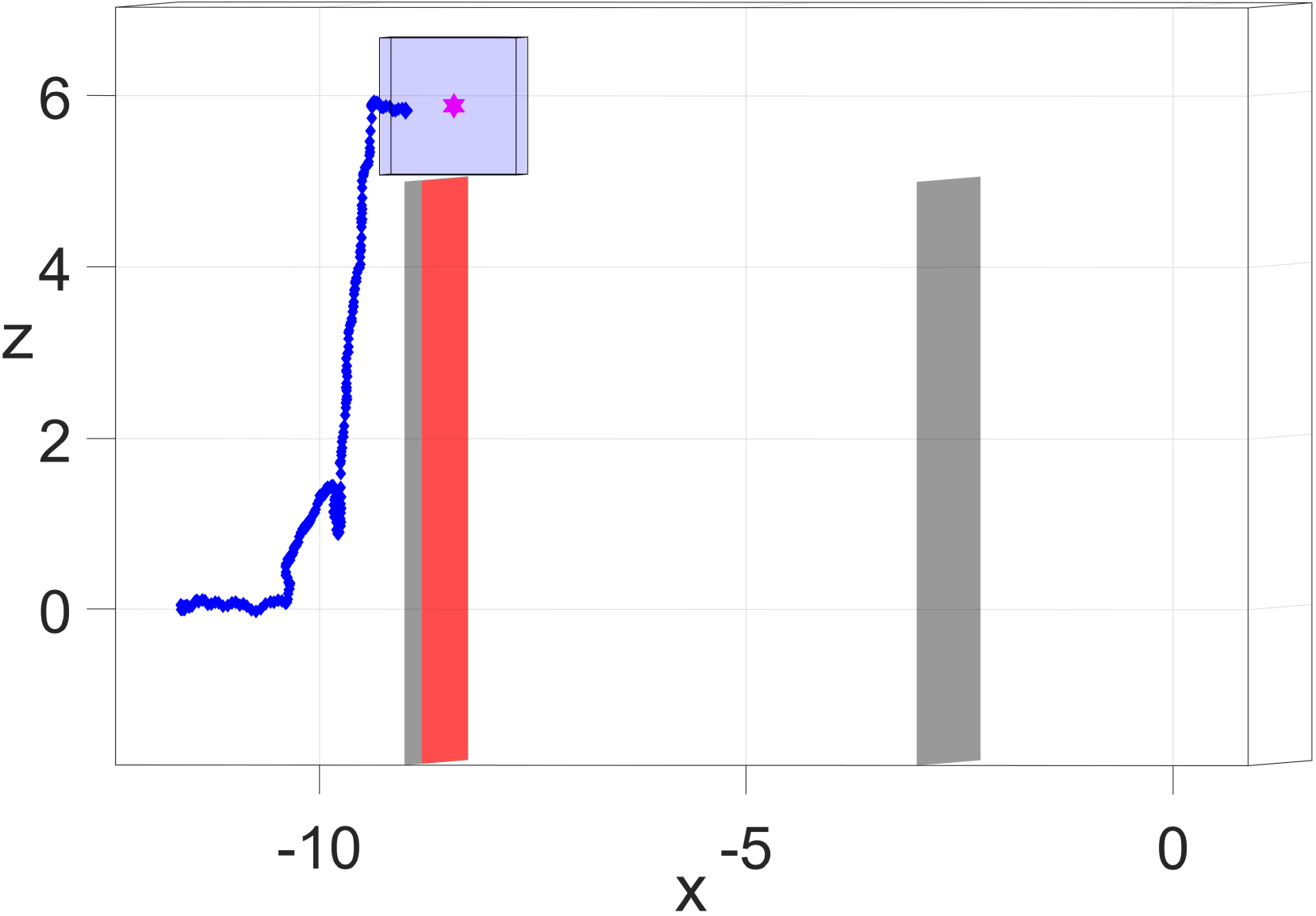}
		\caption{}
	\end{subfigure}
	
	\begin{subfigure}[t]{0.49\columnwidth} \label{subfig:sim_2}
		\includegraphics[width=\columnwidth]{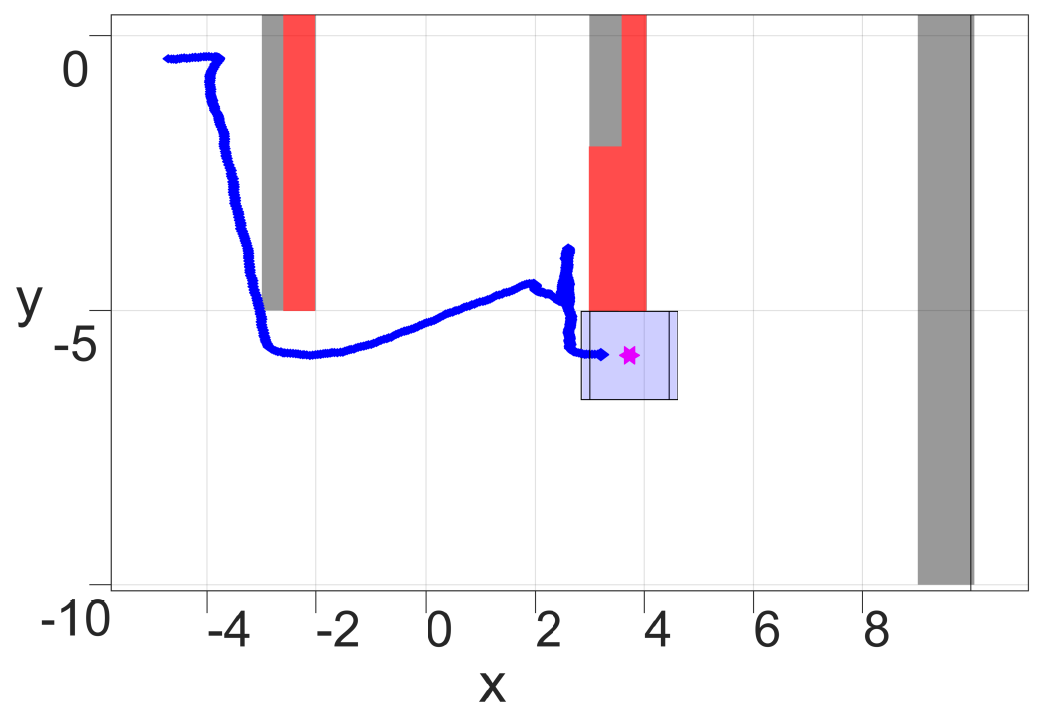}
		\caption{}
	\end{subfigure}  
	\begin{subfigure}[t]{0.49\columnwidth} \label{subfig:sim_3}
		\includegraphics[width=\columnwidth]{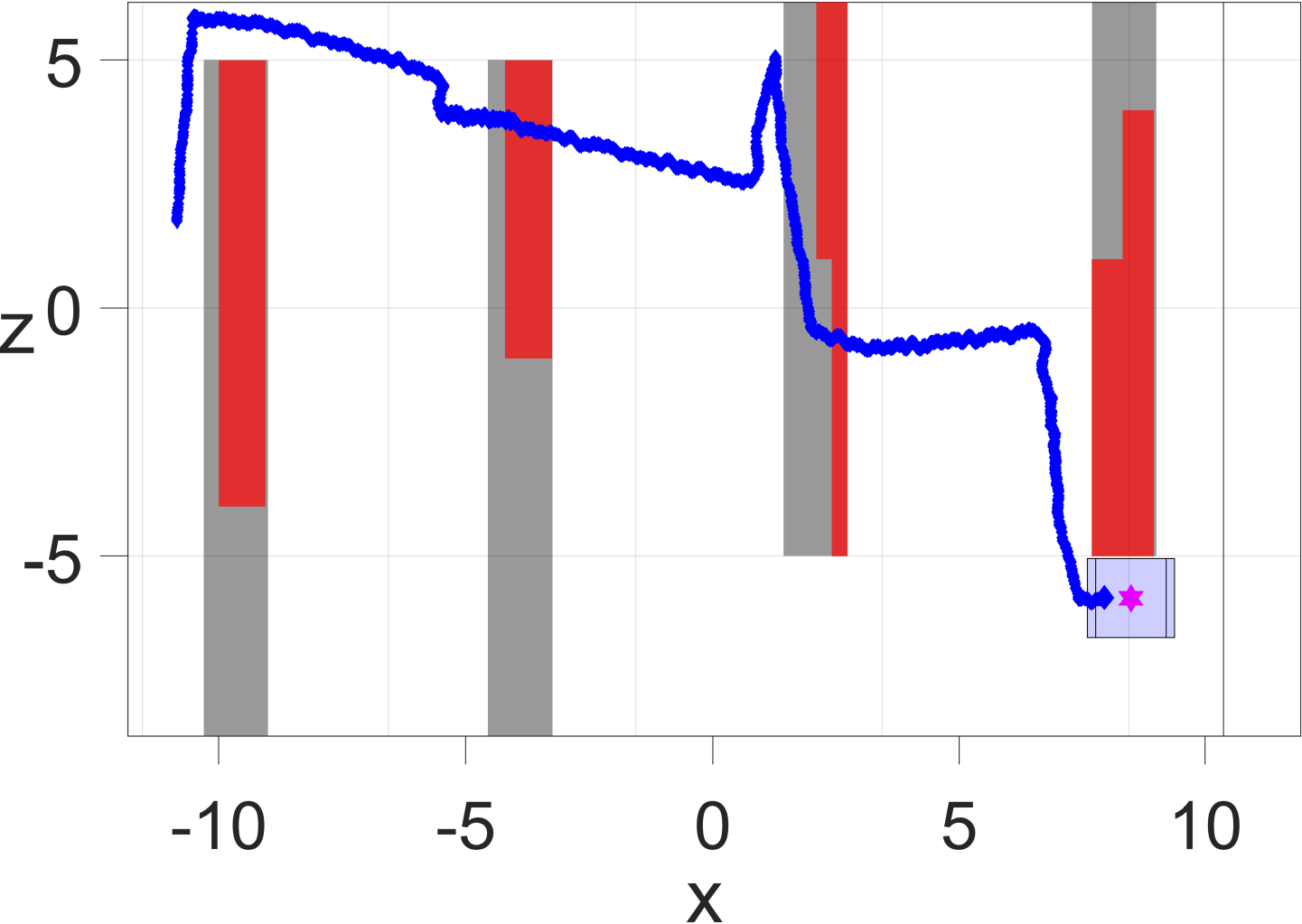}
		\caption{}
	\end{subfigure}
	\vspace{-.1in}
	\caption{Numerical simulation. The tracking model trajectory is shown in blue, the planning model position in magenta, unseen obstacles in gray, and seen obstacles in red. The translucent blue box represents the tracking error bound. The top left subplot shows the entire trajectory; the other subplots zoom in on the positions marked in the top left subplot. The camera angle is also adjusted to illustrate our theoretical guarantees on tracking error and robustness in planning. A video of this simulation can be found at https://youtu.be/ZVvyeK-a62E \label{fig:sim}}
	\vspace{-.2in}
\end{figure}

Fig. \ref{fig:sim} shows the simulation results. Four time snapshots are shown. The initial position is $(-12, 0, 0)$, and the goal position is $(12, 0, 0)$. The top left subplot shows the entire trajectory from beginning to end. In all plots, a magenta star represents the position of the planning model; its movement is based on the paths planned by RRT, and is modeled by a 3D holonomic vehicle with a maximum speed. The blue box around the magenta star represents the tracking error bound.
The position of the tracking model is shown in blue. Throughout the simulation, the tracking model's position is always inside the tracking error, in agreement with Proposition \ref{prop:main}. In addition, the tracking error bound never intersects with the obstacles, a consequence of the RRT planner planning with respect to a set of augmented obstacles (not shown). In the latter two subplots, one can see that the quadrotor appears to be exploring the environment briefly before reaching the goal. In this paper, we did not employ any exploration algorithm; this exploration behavior is simply emerging from replanning using RRT whenever a new part (a $3$ m$^2$ portion) of an obstacle is sensed.

\section{Conclusions and Future work}
In this paper we introduced our new tool FaSTrack: Fast and Safe Tracking. This tool can be used to add robustness to various path and trajectory planners without sacrificing fast online computation. So far this tool can be applied to unknown environments with a limited sensing range and static obstacles. We are excited to explore several future directions for FaSTrack in the near future, including exploring robustness for moving obstacles, adaptable error bounds based on external disturbances, and demonstration on a variety of planners.
\vspace{-.1in}


\bibliographystyle{IEEEtran}
\bibliography{references}
\end{document}